\documentclass[sigconf]{acmart}

\usepackage{algorithm}
\usepackage{algorithmic}
\usepackage{multirow}
\usepackage{tabularx}
\usepackage{subfigure}
\usepackage{tcolorbox}
\usepackage{array}
\usepackage{longtable}
\usepackage{booktabs}
\usepackage{enumitem} 
\newcommand{\mynl}{\textcolor{gray}{\textbackslash n}}
\AtBeginDocument{%
  }

\setcopyright{acmlicensed}
\copyrightyear{2018}
\acmYear{2018}
\acmDOI{XXXXXXX.XXXXXXX}
\acmConference[Conference acronym 'XX]{Make sure to enter the correct
  conference title from your rights confirmation email}{June 03--05,
  2018}{Woodstock, NY}
\acmISBN{978-1-4503-XXXX-X/2018/06}

\begin{document}

\title{Data-centric Federated Graph Learning with Large Language Models}

\author{Bo Yan}
\affiliation{%
  \institution{Beijing University Of Posts and Telecommunications}
  \city{Beijing}
  \country{China}
}

\author{Zhongjian Zhang}
\affiliation{%
  \institution{Beijing University Of Posts and Telecommunications}
  \city{Beijing}
  \country{China}
}

\author{Huabin Sun}
\affiliation{%
  \institution{Beijing University Of Posts and Telecommunications}
  \city{Beijing}
  \country{China}
}

\author{Mengmei Zhang}
\affiliation{%
  \institution{China Telecom Bestpay}
  \city{Beijing}
  \country{China}
}

\author{Yang Cao}
\affiliation{%
  \institution{Institute of Science Tokyo}
  \city{Tokyo}
  \country{Japan}
}

\author{Chuan shi}
\authornote{Corresponding author}
\affiliation{%
  \institution{Beijing University Of Posts and Telecommunications}
  \city{Beijing}
  \country{China}
}


\renewcommand{\shortauthors}{Trovato et al.}

\begin{abstract}
In federated graph learning (FGL), a complete graph is divided into multiple subgraphs stored in each client due to privacy concerns, and all clients jointly train a global graph model by only transmitting model parameters. A pain point of FGL is the heterogeneity problem, where nodes or structures present non-IID properties among clients (e.g., different node label distributions), dramatically undermining the convergence and performance of FGL. To address this, existing efforts focus on design strategies at the model level, i.e., they design models to extract common knowledge to mitigate heterogeneity. However, these model-level strategies can’t fundamentally address the heterogeneity problem as the model needs to be designed from scratch when transferring to other tasks. 
Motivated by large language models (LLMs) having achieved remarkable success, we aim to utilize LLMs to fully understand and augment local text-attributed graphs, to address data heterogeneity at the data level. In this paper, we propose a general framework LLM4FGL that innovatively decomposes the task of LLM for FGL into two sub-tasks theoretically. Specifically, for each client, it first utilizes the LLM to generate missing neighbors and then infers connections between generated nodes and raw nodes.
To improve the quality of generated nodes, we design a novel federated generation-and-reflection mechanism for LLMs, without the need to modify the parameters of the LLM but relying solely on the collective feedback from all clients. After neighbor generation, all the clients utilize a pre-trained edge predictor to infer the missing edges. Furthermore, our framework can seamlessly integrate as a plug-in with existing FGL methods. Experiments on three real-world datasets demonstrate the superiority of our method compared to advanced baselines.
\end{abstract}






\maketitle

\section{Introduction}
Graph data, such as social networks, e-commerce networks, and protein interaction networks, is ubiquitous in the real world. As a typical technique to model graph data, graph neural networks (GNNs) utilize message passing to learn the relations between nodes and labels, having made great progress in many applications, e.g., recommender systems and biomedicines. Traditional GNNs hold a basic assumption that the graph data is centrally stored, which is unrealistic in some privacy-sensitive scenarios, such as financial systems and hospitals. In these cases, the whole graph is divided into multiple subgraphs owned by each data holder (clients). In recent years, federated graph learning (FGL) enables all clients to collaboratively train a global GNN without data exposure. 

A key challenge of FGL is data heterogeneity among different clients. Roughly speaking, there are two kinds of heterogeneity issues in the current FGL. (1) Node heterogeneity, where different clients own different node label distributions. (2) Edge heterogeneity, where different clients may have different subgraph structures, e.g., different communities. Current methods mostly focus on model-level design to alleviate these heterogeneity issues \cite{DBLP:conf/icde/LiWZSLW24,DBLP:conf/kdd/FuCZ0L24,DBLP:conf/ijcai/ZhuLWWHL24}. They design models to distill reliable knowledge from local clients \cite{DBLP:conf/ijcai/ZhuLWWHL24}, learn local class-wise structure proxies \cite{DBLP:conf/kdd/FuCZ0L24}, or perform homophilous/heterophilous propagations \cite{DBLP:conf/icde/LiWZSLW24}, etc. We argue that these model-level strategies could not fundamentally address the heterogeneity issue, as the inherent data heterogeneity still exists and may also affect the performance when transferring the model to other tasks.

Motivated by this limitation, a natural and fundamental solution to tackle data heterogeneity in FGL is designing strategies from a data-centric view, i.e., directly generating local data to mitigate the distribution gap among clients. Recently, Large language models (LLMs), have demonstrated powerful abilities in understanding and generating complex text, achieving remarkable success in the fields of natural language processing \cite{DBLP:journals/csur/MinRSVNSAHR24}, computer vision \cite{DBLP:conf/iclr/Zhu0SLE24}, and text-attributed graphs (TAGs)~\cite{DBLP:conf/iclr/HeB0PLH24}. Empowered by LLMs, the performance of various graph tasks (e.g., node classification and adversarial robustness) has largely improved.
However, their potential benefits when applied to FGL remain largely uncharted.
Considering this, we aim to unleash the power of LLMs for FGL and propose to leverage LLMs to generate local graph data. 

Thus, the pivotal challenge is to improve the quality of generated data by LLMs for FGL, which is further compounded by the privacy-sensitive scenarios in the federated setting. Firstly, generating data to mitigate heterogeneity requires the LLMs to know the whole data distribution.  However, the LLMs are generally deployed as trustworthy online services (e.g., GPT-4) or local open-sourced models (e.g., Gemma), and due to privacy concerns, they can only observe the local data of specific clients but can not access the whole data of all clients. As a result, It’s difficult for LLMs to infer the whole data distribution and generate high-quality data. Secondly, LLMs present superiority in their text data understanding and generation ability but are still under-explored in generating graph data. Very few studies focus on generating graph data from scratch \cite{DBLP:journals/corr/abs-2403-14358}\cite{DBLP:journals/corr/abs-2410-09824}. In contrast, we aim to generate the unseen data based on the observed partial data, which is totally different from existing LLM-based simple graph generation. 

To address these challenges, in this paper, we investigate the potential of LLMs for data-centric FGL. We propose a general framework LLM4FGL to mitigate the heterogeneity issue from a data-centric view. Concretely, we first decompose the LLM-based generation task into two sub-tasks theoretically, which generate node text based on LLM and then infer the missing edges between original nodes and newly generated nodes. In this way, LLM can focus on tasks they excel at (i.e., text understanding and generation), with no need to handle structure inference. For the LLM-based generation task, we meticulously design a generation-and-refection mechanism to improve the quality of generated textual data. All the clients first generate textual neighbor nodes locally based on the prompts including the current node and its neighbors. Then they collaboratively train a global GNN mode based on augmented local data. To quantify the quality of generated data, we utilize the prediction confidence produced by the GNN model to be the collective feedback. The nodes with low confidence will be informed to LLMs and LLMs will regenerate neighbors based on previous generated neighbors. To further obtain high-quality new edges, we propose collaboratively training a lightweight edge predictor to infer the edges between generated nodes and original nodes. The edge predictor is trained by clients' local edges, and the top-k most possible edges are selected as new edges. 
The reflection of generated nodes and inferring edges are combined during each round of federated training. The final generated graph can be directly utilized to train simple GNN models without bothering model-level designs. In this regard, our framework can be seen as a plug-in from the data level and can be seamlessly integrated into existing model-level FGL methods and further improve their performance. 

In summary, our contributions lie in three aspects: 

(1) To the best of our knowledge, we are the first to utilize LLMs for FGL, broadening the application scope of LLMs and also significantly advancing the progress of FGL. We propose a general framework called LLM4FGL to tackle the heterogeneity issue of FGL from the data-centric view. LLM4FGL decomposes the task of tackling heterogeneity into two sub-tasks theoretically, which first generate high-quality textural node data by LLMs, and then infer the edges between original nodes and generated nodes.

(2) Building upon high-level principles by the general framework, we propose a generation-and-refection mechanism to guide the LLM to iteratively generate node texts based on reflection on the previous generations. Then a lightweight edge predictor is jointly trained by all the clients to infer edges between original nodes and generated nodes. Moreover, our framework can be seamlessly integrated into existing model-level methods as a data-level augmentation. 

(3) Experiments on three datasets manifest the superiority of our framework over existing FGL counterparts. Furthermore, as a plug-in module, our method can also largely improve existing methods.

\section{Related work}
\subsection{Federated learning on graph}


FGL applies the principles of FL to graph data, enabling multiple clients to collaboratively learn a global model with distributed data while protecting data privacy. A key challenge in FGL, similar to generic FL, is data heterogeneity, which includes both node heterogeneity and edge heterogeneity. Several approaches have been proposed to tackle this issue. FedSage+ \cite{DBLP:conf/nips/ZhangYLSY21} addresses heterogeneity caused by missing connections by training a linear generator to generate missing neighbor node information. However, the process of information sharing increases the communication overhead between clients. Therefore, Fed-PUB \cite{DBLP:conf/icml/BaekJJYH23} takes the perspective of subgraph communities and selectively updates subgraph parameters by measuring subgraph similarities. FedGTA \cite{DBLP:journals/corr/abs-2401-11755} enhances model aggregation by utilizing topology-related smoothing confidence and graph moments. These methods, however, may be misled by unreliable class-wise knowledge during server-side model aggregation. So FedTAD \cite{DBLP:conf/ijcai/ZhuLWWHL24} quantifies class confidence and employs distillation to extract reliable knowledge from individual clients into the global model. Additionally, some methods, such as AdaFGL \cite{DBLP:conf/icde/LiWZSLW24}, adopt personalization by combining a global federated knowledge extractor with local clients for personalized optimization. However, existing strategies primarily focus on the model level, which cannot fully address the heterogeneity caused by data variability. In contrast, we aim to leverage the powerful understanding and generation capabilities of LLMs to tackle this issue at the data level.

\subsection{LLMs for graph learning}
In recent years, Large Language Models (LLMs) have been widely used in various graph learning tasks, to surpass traditional GNN-based methods and yield state-of-the-art performance. Here, we categorize different methods based on the specific roles of LLMs in tasks. Specifically, LLM-as-enhancer\cite{Xie2023GraphAwareLM,He2023HarnessingEL,Liu2023OneFA,Tan2023WalkLMAU,He2024UniGraphLA} methods enhance node features by encoding or generating additional text using an LLM, while LLM-as-predictor\cite{Wang2023CanLM,Zhao2023GIMLETAU,Chen2023ExploringTP,Zhang2024GraphTranslatorAG,Tang2024HiGPTHG} methods serialize graphs into natural language to feed into an LLM for direct reasoning. Additionally, some methods also utilize LLMs as defenders for adversarial attacks\cite{Zhang2024CanLL}, attackers for graph injection attacks\cite{Lei2024IntrudingWW}, annotators for supervised learning\cite{Chen2023LabelfreeNC}, generators\cite{DBLP:journals/corr/abs-2403-14358,DBLP:journals/corr/abs-2410-09824,Yu2023LeveragingLL}, controllers\cite{Wang2023GraphNA}, and task planners\cite{wu2024can}. 
Although the above methods achieve satisfactory performance in graph learning tasks in non-federated scenarios, the potential of LLMs for FGL remains unexplored. In this paper, we further unleash the power of LLMs in FGL by using them to address the key challenges of node heterogeneity and edge heterogeneity.



\section{Preliminaries}
\subsection{Text-attributed graphs }
A text-attributed graph (TAG) is defined as $\mathcal{G}=(\mathcal{V}, \mathcal{E}, \mathcal{S})$ with adjacency matrix $\mathbf{A}\in\mathbb{R}^{\left | \mathcal{V} \right | \times \left | \mathcal{V} \right |}$ and node-level textual information $\mathcal{S} = \{\mathbf{s}_{1},\dots,\mathbf{s}_{\left | \mathcal{V} \right |}\}$, where $\mathcal{V}=\{v_1,\ldots, v_{\left | \mathcal{V} \right |}\}$ is the node set. $\mathbf{A}$ contains topological information of the graph and $\mathbf{A}_{ij}=1$ denotes the nodes $v_i$ and $v_j$ are connected, otherwise $\mathcal{A}_{ij}=0$. In this work, we focus on the node classification task on TAGs, where each node $v_i$ corresponds to a category label $y_i$. The textural information $\mathcal{S}$ is usually encoded by some embedding techniques ~\cite{mikolov2013distributed, DBLP:conf/emnlp/ReimersG19} as the node feature matrix $\mathbf{X}=\{\mathbf{x}_{1},\dots,\mathbf{x}_{\left | \mathcal{V} \right |}\}$ to train GNNs, where $\mathbf{x}_i \in \mathbb{R}^d$. Given some labeled nodes $\mathcal{V}^L \subset \mathcal{V}$, the goal is to train a GNN $f(\mathbf{A}, \mathbf{X})$ to predict the labels of the remaining unlabeled nodes $\mathcal{V}^U=\mathcal{V}\setminus \mathcal{V}^L$.

\subsection{Federated graph learning }
Given a set of local graphs $\{\mathcal{G}_1,\mathcal{G}_2,\dots,\mathcal{G}_n\}$ stored in the corresponding client, federated graph learning (FGL) aims to collaboratively train a global GNN model $f_{\theta}$ by only exchanging intermediate parameters \cite{DBLP:conf/nips/ZhangYLSY21, DBLP:journals/corr/abs-2406-18937} and utilizes $f_{\theta}$ to predict the unlabeled nodes $\mathcal{V}^U_i=\mathcal{V}_i\setminus \mathcal{V}_i^L$ in each client. Typically, at each communication round $t$ of FGL, each client first initializes local parameter $\theta_i^t$ with the global model parameters $\theta^t$, then performs local training with local data to obtain the model gradients $\mathbf{g}_i^t$. All the gradients are uploaded to the server for aggregation. By employing FedAvg \cite{DBLP:conf/aistats/McMahanMRHA17}, the server aggregates gradients with:

\begin{equation} 
    \mathbf{g}^{t+1} = \sum_{i=1}^{n}\frac{|\mathcal{V}_i|}{\sum_{j=1}^n |\mathcal{V}_j|}\mathbf{g}_i^t,
\label{eq:agg}
\end{equation}
Finally, the server updates the global model parameter with $\mathbf{g}^{t+1}$.

\begin{table}[ht]
    \centering
    \caption{Notations.}
    \begin{tabular}{>{\centering\arraybackslash}p{1.4cm} | p{6cm}}
        \toprule
        \multicolumn{1}{c|}{\textbf{Symbol}} & \multicolumn{1}{c}{\textbf{Description}} \\
        \midrule
        $\mathcal{G} = (\mathcal{V},\mathcal{E}, \mathcal{S})$ & A graph with node set $\mathcal{V}$, edge set $\mathcal{E}$, and node textural information set $\mathcal{S}$. \\
        $\mathcal{G}^o_i$ & Original graph of client $i$\\
        $\mathcal{G}^*_i$ & Augmented graph of client $i$\\
        $\mathbf{A}$ & Adjacency matrix of the graph. \\
        $\mathbf{X}$ & Node feature matrix. \\
        $y_i$ & Label of node $v_i$. \\
        $f_\theta$ & Graph neural network (GNN) model with parameters $\theta$. \\
        $\text{LLM}_i$ & $i$-th Large Language Model used in client $i$. \\
        $\mathcal{G}_i$ & Local subgraph stored in client $i$. \\
        $\mathcal{L}$ & Loss function for training the model. \\
        $p_i$ & Prediction confidence of node $v_i$. \\
        \bottomrule
    \end{tabular}
    \label{tab:notation}
\vspace{-3mm}
\end{table}

\section{Methodology}

\begin{figure*}[h]
    \centering
    \includegraphics[scale=.69]{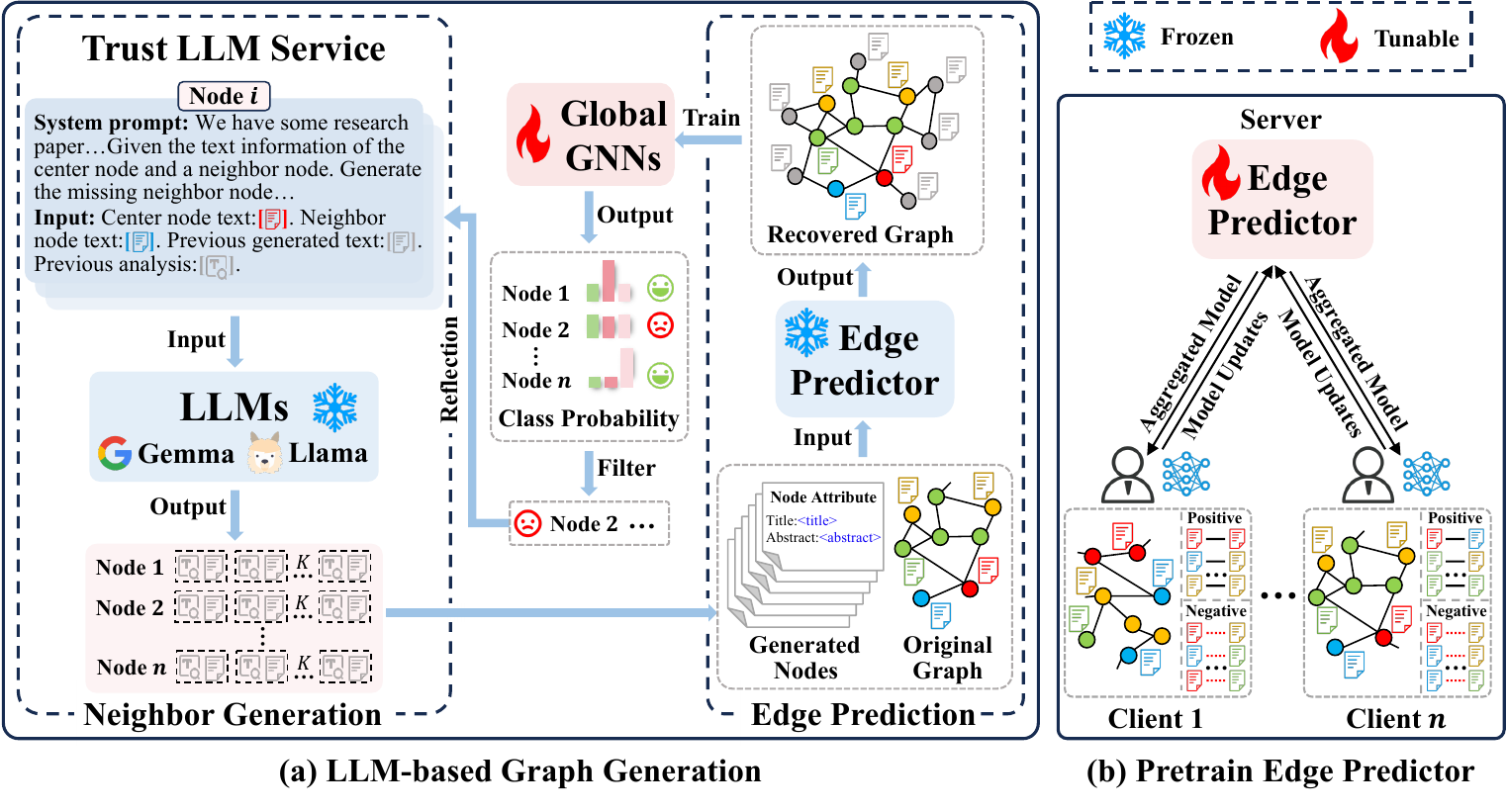}
    \caption{The overall framework of LLM4FGL.}
    \label{fig:framework}
\end{figure*}

In this section, we present a detailed description of LLM4FGL. We first introduce the theoretical framework of LLM for FGL. Following this, we delve into the LLM-based graph data generation process. Then we describe the edge prediction mechanism. Finally, we summarize the overall algorithm for LLM4FGL. An overview of the proposed framework is shown in Figure~\ref{fig:framework}, and the notations used in this paper are reported in Table~\ref{tab:notation}.

\subsection{A theoretical framework of LLM for FGL}
Assume an original local graph $\mathcal{G}^o=(\mathcal{V}^o, \mathcal{E}^o, \mathcal{S}^o)$ in one client, to tackle the issue of heterogeneity in FGL, an ideal solution from the data level is to augment $\mathcal{G}^o$ so that the global GNN model can be trained over identical augmented graph $\mathcal{G}_*=(\mathcal{V}^*, \mathcal{E}^*, \mathcal{S}^*)$ without any heterogeneity. Here, $\mathcal{E}^*=(\mathcal{E}^g, \mathcal{E}^o, \mathcal{E}^{go})$ including the edge set $\mathcal{E}^g$ between generated nodes, $\mathcal{E}^o$ between original nodes, and $\mathcal{E}^{go}$ between generated nodes and original nodes. $\mathcal{S}^*=(\mathcal{S}^o, \mathcal{S}^g)$ includes original node texts $\mathcal{S}^o$ and generated $m$ node texts $\mathcal{S}^g=(s^g_1, s^g_2, \cdots, s^g_m)$. In this way, the task can be formulated as modeling a conditional probability $P(\mathcal{G}^*|\mathcal{G}^o)$ and can be further decomposed into:

\begin{equation}
\begin{aligned}
P(\mathcal{G}^*|\mathcal{G}^o)
&=P(\mathcal{V}^*,\mathcal{E}^*, \mathcal{S}^*|\mathcal{G}^o)\\
&=P(\mathcal{E}^o, \mathcal{E}^g, \mathcal{E}^{go}, \mathcal{V}^o, \mathcal{V}^{g},\mathcal{S}^g, \mathcal{S}^o|\mathcal{V}^o,\mathcal{E}^o, \mathcal{S}^o)\\
&=P(\mathcal{E}^g, \mathcal{E}^{go}, \mathcal{V}^{g},\mathcal{S}^g|\mathcal{V}^o, \mathcal{E}^o, \mathcal{S}^o)\\
&=P(\mathcal{S}^g, \mathcal{V}^g|\mathcal{G}^o)P(\mathcal{E}^+|\mathcal{S}^g, \mathcal{V}^g, \mathcal{G}^o).
\end{aligned}
\label{eq:objective}
\end{equation}
Here we use $\mathcal{E}^+=(\mathcal{E}^{g}, \mathcal{E}^{go})$ to denote all the generated structures. 

From Eq. (\ref{eq:objective}) we can see that the objective can be divided into two conditional distributions and thus the task is divided into two sub-tasks: (1) Utilizing the original local graph $\mathcal{G}$ to generate some textural nodes ($\mathcal{S}^g$, $\mathcal{V}^g$), and (2) Utilizing ($\mathcal{S}^g$, $\mathcal{V}^g$) and $\mathcal{G}_o$ to infer the newly added graph structures $\mathcal{E}^+$. To facilitate modeling the two distributions, we first give two basic assumptions. 

\noindent \textbf{Assumption 1}. Given the original graph, all the generated nodes have independent-and-identically (I.I.D) distributions.

Assumption 1 is a common assumption in classic machine learning theories, where different samples are generated or sampled from the same distribution independently.

\noindent \textbf{Assumption 2}. Each generated node only depends on one node and its $l$-hop neighbors in the original graph.

Assumption 2 stems from a common assumption in the graph domain that the label of a node is usually determined by its $l$-hop neighbors. Motivated by this, we assume that the generated node also depends on the original graph's one $l$-hop subgraph.

With these two assumptions, we can further decompose the first conditional distributions in Eq. (\ref{eq:objective}). For $P(\mathcal{S}^g, \mathcal{V}^g|\mathcal{G}^o)$, we have:

\begin{equation}
\begin{aligned}
P(\mathcal{S}^g, \mathcal{V}^g|\mathcal{G}^o)
&=P(s^g_1, v^g_1|\mathcal{G}^o)P(s^g_2, v^g_2|\mathcal{G}^o) \cdots P(s^g_m, v^g_m|\mathcal{G}^o)\\
&=\prod_{k=1}^{m}P(s_k^g, v^g_k|\mathcal{G}^o)\\
&=\prod_{k=1}^{m}P(s_k^g, v^g_k|\{s_i, v_i\}_{v_i \in \mathcal{N}^l_{j_k}},s_{j_k},v_{j_k}),
\end{aligned}
\label{eq:ob1_temp}
\end{equation}
where $\mathcal{N}^l_{j_k}$ denotes the $l$-hop neighbors of node $v_{j_k}$. Until now, we just need to model the two conditional distributions described above. However, since we can only observe local data in the federated setting, it's unrealistic to augment local data to approximate the global data distribution without any other prior knowledge. Fortunately, the global model $F_{\theta}=\{f_{\theta_1},f_{\theta_2},\cdots,f_{\theta_T}\}$ trained by all clients' local data is accessible during each communication round $t$ in the federated training. We can further modify the two distributions into:
\begin{equation}
\begin{aligned}
P(\mathcal{S}^g, \mathcal{V}^g|\mathcal{G}^o)=
\prod_{k=1}^{m}P(s_k^g, v^g_k|\{s_i, v_i\}_{v_i \in \mathcal{N}^l_{j_k}},s_{j_k},v_{j_k}, F_\theta)
\end{aligned}
\label{eq:ob1}
\end{equation}
and
\begin{equation}
\begin{aligned}
P(\mathcal{E}^+|\mathcal{S}^g, \mathcal{V}^g, \mathcal{G}^o) = P(\mathcal{E}^g, \mathcal{E}^{go}|\mathcal{S}^g, \mathcal{V}^g, \mathcal{G}^o, F_{\theta}).
\end{aligned}
\label{eq:ob2}
\end{equation}
Eq. (\ref{eq:ob1}) and Eq. (\ref{eq:ob2}) give a high-level theoretical guideline to conduct LLM-based FGL, which decomposes the complex objective into two independent sub-tasks. They also facilitate LLM to handle tasks (i.e., understand and generate textural data) they excel in and leave challenging tasks (i.e., infer graph structure) to other modules.

\subsection{LLM-based graph generation}
To model the first conditional distribution depicted in Eq. (\ref{eq:ob1}), we leverage LLMs to directly generate node texts. As shown in Figure~\ref{fig:framework} (a), we assume the LLM is deployed as a trustworthy online service (e.g., GPT-4), and design the following prompt templates to generate the missing neighbors for each node $v_i$:

\begin{tcolorbox}[top=2pt, bottom=2pt, left=4pt, right=4pt]
\textbf{System prompt:} We have some research paper topics: 
\{\textcolor{blue}{list of categories}\}. Given the text information of the center node and a neighbor node, please analyze the research topic to which the center node belongs (approximately 100 words), and generate a missing neighbor that logically complements both the center node (including title and abstract). Your response should be in JSON format with only two keys: "Topic Analysis" for the topic analysis of the center node, and "Missing Neighbor" for the generation of your node. Note that the generated node only has two keys: "title" and "abstract". Do not discuss anything else. Please provide responses in JSON format. Your answer must be in JSON format.\\
\textbf{User content:} Center Node $v_i$$\rightarrow$\{\textcolor{blue}{Title, Abstract}\}.\mynl\mynl Neighbor Node $v_j$$\rightarrow$\{\textcolor{blue}{Title, Abstract}\}.
\end{tcolorbox}

In the prompt templates, we let LLMs generate complementary information to the center node's text, thus truly understanding the center node's topic (category) is very important. Motivated by \cite{Zhang2024CanLL}, we also let LLM generate an "Analysis" to facilitate an inference process in LLMs regarding prediction. Guided by Eq. (\ref{eq:ob1}), we just need to feed one node text $s_i$ and its $l$-hop neighbors into LLMs. However, LLMs have limitations in token length and too many neighbors may bring noise, leading to reduce the LLMs' confidence in determining the node category. Consider this, we only feed one 1-hop neighbor $v_j$'s text $s_j$ into LLMs in each query and query the LLM $n_i$ times for one node to obtain $n_i$ new neighbors. Note that $n_i$ may be larger than the actual number of neighbors and we perform neighbor sampling to generate each prompt. In this way, the LLMs can not only fully consider neighbor information to understand the center node's topic but also avoid noise caused by multiple neighbors.

Through the above-designed prompt, each client can augment local graph data independently. Although this may improve the local model performance, it still suffers from heterogeneity issues when jointly training a global model by all the clients, since LLMs do not know other clients' data distribution. To make the generated nodes toward the global data distribution, we further propose a generation-and-reflection mechanism. Motivated by \cite{DBLP:conf/nips/WangLSY21, DBLP:conf/icml/GuoPSW17} that the probability (confidence) of the predicted class label should reflect its ground truth correctness likelihood, we utilize confidences of nodes to guide the LLM to reflect generated nodes. Concretely, at each reflection step $t$, the confidence score of a node $v_i$ is obtained from the global GNN $f_{{\theta}_t}$ as follows:

\begin{equation}
\begin{aligned}
&\hat{p_i} = \max_c \hat{z}_{i,c}=\max_c \{\hat{z}_{i,1},\cdots,\hat{z}_{i,C}\}\\
&\mathbf{Z} = [\mathbf{z}_1,\cdots,\mathbf{z}_{|\mathcal{V}|}]^{\mathrm{T}} = f_{\theta}(\mathbf{A},\mathbf{X}),
\end{aligned}
\label{eq:confidence}
\end{equation}
where $C$ is the number of node categories, $\mathbf{z}_i=[z_{i,1},\cdots,z_{i,C}]$ and $\hat{z}_{i,c} = \frac{exp(z_{i,c})}{\sum_{j=1}^{C}exp(z_{i,j})}$ is the softmax operation. After obtaining all local nodes' prediction confidence $\hat{\mathbf{P}}=\{\hat{p_1},\cdots,\hat{p}_{|\mathcal{V}|}\}$, each client will select the top-$k$ nodes with the lowest confidence to feed into LLMs for reflection. Specifically, the reflection prompt is designed as follows.

\begin{tcolorbox}[top=2pt, bottom=2pt, left=4pt, right=4pt]
\textbf{System prompt:} We have some research paper topics:
\{\textcolor{blue}{list of categories}\}. Given the text information of the center node and a neighbor node, please analyze the research topic to which the center node belongs and generate a missing neighbor that logically complements both the center node (including title and abstract). This is an iterative self-reflection process, where the quality of generated neighbor is continuously optimized based on feedback. We will provide previously generated possibly wrong analysis and neighbors. You should reflect potential reasons for their low quality (You may have misjudged the topic of the central node or generated neighbors for the wrong topic). Please regenerate a new neighbor and a new analysis of center nodes's topic, and your reflection. Your response should be in JSON format with only Three keys: "Topic Analysis" for the topic analysis of the center node, "reflection" for the reflection of your previous analysis, and "Missing Neighbor" for the generation of your node. Note that the generated node only has two keys: "title" and "abstract". Do not discuss anything else. Please provide response in JSON format. Your answer must be in JSON format. \\
\textbf{User content:} Center Node $v_i$$\rightarrow$ \{\textcolor{blue}{Title, Abstract}\}.\mynl\mynl Neighbor Node $v_j$$\rightarrow$ \{\textcolor{blue}{Title, Abstract}\}.\mynl\mynl Previously generated neighbor $v^p_j$$\rightarrow$ \{\textcolor{blue}{Title, Abstract}\} \mynl\mynl Previous analysis $S^p_j$$\rightarrow$ \{\textcolor{blue}{Analysis}\}.
\end{tcolorbox}

For each node, we feed the LLM with the previously generated neighbor node and previous analysis. A reflection instruction is also attached to let LLM perform self-reflection based on previously generated textural information. Finally, LLM will generate a new neighbor which may be better for helping predict the center node's label. The reflection is an iterative process and the LLM will gradually generate nodes to help improve the global model's prediction confidence. This process also follows the theoretical guidance in Eq. (\ref{eq:ob1}), where we utilize other clients' knowledge through a collaboratively trained global model to guide the LLM-based generation but keep local private data locally.

\subsection{Edge prediction}
Given the generated node texts by LLMs, we aim to infer the underlying structures of all nodes. Guided by the second conditional distribution depicted in Eq. (\ref{eq:ob2}), we propose to collaboratively train an edge predictor $f_{\phi}$ to infer the structure $\mathcal{E}^+=(\mathcal{E}^g, \mathcal{E}^{go})$. Considering the computation and storage limitations of local clients, we select a lightweight MLP as an edge predictor. As depicted in Figure~\ref{fig:framework} (b), for each client, if the edge between $v_i$ and $v_j$ exists (i.e., $\mathbf{A}_{i,j}=1$), it will be selected as a positive edge, and all the existing edges consist of positive edge set $\mathcal{E}^p$. We sample the equal number of negative edges (i.e., $\mathcal{A}_{i,j}=0)$ to form the negative edge set $\mathcal{E}^n$. Next, we concatenate $\mathbf{x}_i$ and $\mathbf{x}_j$ to obtain the edge representation and feed it into the edge predictor to obtain the edge probability $\hat{y}_e(v_i,v_j) = \text{MLP}(\mathbf{x}_i \Vert \mathbf{x}_j$). The cross-entropy loss is used to optimize the parameters:

\begin{equation}
\mathcal{L}_{ep} = - \frac{1}{|\mathcal{E}^p \cup \mathcal{E}^n|} \sum_{i=1}^{|\mathcal{E}^p \cup \mathcal{E}^n|}  y_i \log(\hat{y}_i) + (1 - y_i) \log(1 - \hat{y}_i).
\label{eq:ep_loss}
\end{equation}
Finally, each client trains the edge predictor locally and upload gradients of parameters to the server for aggregation. Note that the training of the edge predictor is decoupled with the LLM-based generation, so it can be trained in advance and then as a callable frozen model.

After the edge predictor is well trained, we can infer the structure $\mathcal{E}^+=(\mathcal{E}^g, \mathcal{E}^{go})$. For each possible edge between node $v_i$ and $v_j$ within $\mathcal{E}^+$, we feed the edge representation into $f_{\phi}$ to obtain the probability of edge existence. Then we select the most possible top-$k$ edges as the newly added edges. Note that since we generate nodes based on one node in the original graph, we directly add edges between the generated nodes and the original node.

\subsection{Overall algorithm}
The overall algorithm of LLM4FGL is depicted in Algorithm \ref{alg}. Firstly, all the clients first jointly train a global edge predictor $f_{\phi}$. Each client queries its own LLM to obtain the augmented graph. Then, all the clients jointly train a global GNN model $f_{\theta}$. At each communication round $t$, if $t$ is in the refection rounds,  each client leverages $f_{\theta}$ to obtain nodes' confidence score and selects top-$k$ nodes with the lowest confidence for refection with LLMs. The newly generated nodes are connected to the currently augmented graph through $f_{\phi}$. Finally, the final augmented graph can be used for downstream tasks.

\begin{algorithm}
    \renewcommand{\algorithmicrequire}{\textbf{Input:}}
    \renewcommand{\algorithmicensure}{\textbf{Output:}}
    \caption{LLM4FGL} 
    \label{alg}

    \begin{algorithmic}[1]
        \REQUIRE Local subgraphs $\{\mathcal{G}_i^o\}_{i=1}^n$,
        positive edge set $\{\mathcal{E}^p_i\}_{i=1}^n$,
        trustworthy LLMs $\{\text{LLM}_i\}_{i=1}^n$; refection round $\mathcal{R}$
        \ENSURE Augmented subgraphs $\{\mathcal{G}^*_i\}_{i=1}^n$
        
        \STATE Each client $i$ samples negative edge set $\mathcal{E}^n_i$
        \STATE Train a global edge predictor $f_{\phi}$ based on $\{\mathcal{E}^p_i\}_{i=1}^n$ and $\{\mathcal{E}^n_i\}_{i=1}^n$ using Eq. (\ref{eq:ep_loss})
        \STATE Each client $i$ queries $\text{LLM}_i$ based on $\mathcal{G}_i^o$ to obtain initial $\mathcal{G}^*_i$

        \FOR{each communication round $t=1,\cdots,T$}
            \STATE Update global GNN $f_{\theta}$ based on current $\{\mathcal{G}^*_i\}_{i=1}^n$ 
            \IF{$t \in R$}
                \STATE Each client $i$ obtains nodes' confidence $\hat{\mathbf{P}}_i$ using Eq. (\ref{eq:confidence})
                \STATE Each client $i$ selects top-$k$ nodes with lowest confidence and feed into $\text{LLM}_i$ for reflection
                \STATE $\text{LLM}_i$ generates new node set ($\mathcal{S}^g_i$, $\mathcal{V}^g_i$)  for each client $i$
                \STATE Each client $i$ utilizes $f_{\phi}$ to infer the structure $\mathcal{E}^+_i$ and updates $\mathcal{G}^*_i$
            \ENDIF
        \ENDFOR
        \STATE \textbf{return} $\{\mathcal{G}^*_i\}_{i=1}^n$
        
    \end{algorithmic}
\end{algorithm}

\section{Experiments}
In this section, we aim to answer the following research questions to demonstrate the effectiveness of LLM4FGL:
\begin{itemize}[leftmargin=10pt]
    \item \textbf{RQ1:} Compared to existing methods that address the data heterogeneity problem in FGL, how does the proposed LLM4FGL perform? 
    \item \textbf{RQ2:} How effectively can LLM4FGL as a plug-in module enhance existing FGL models? 
    \item \textbf{RQ3:} How do the key components benefit the performance of the LLM4FGL? 
    \item \textbf{RQ4:} How do the hyperparameters impact the performance of LLM4FGL?
\end{itemize}


\begin{table}[tp]
\caption{Statistics of datasets.}
\label{tab:datasets}
\begin{tabular}{cccccc}
\toprule
\multicolumn{1}{c}{\bf Dataset} & \multicolumn{1}{c}{\bf \#Nodes} & \multicolumn{1}{c}{\bf \#Edges} & \multicolumn{1}{c}{\bf \#Classes} & \multicolumn{1}{c}{\bf Train/Val/Test}  \\ 
\midrule
Cora                         & 2,708                     &  5,429                         & 7                    &  140/500/1000                               \\
CiteSeer                     & 3,186                     & 4,277                       & 6                    &    120/500/1000                             \\
PubMed                         & 19,717                     & 44,338                        & 3                    &60/500/1000                              \\
\bottomrule
\end{tabular}
\end{table}

\subsection{Experimental setup}
\noindent \textbf{Dataset}. Following \cite{DBLP:journals/corr/abs-2406-18937}, we perform experiments on three widely used citation network datasets: Cora, CiteSeer, and PubMed. We utilize the processed data by \cite{Chen2023ExploringTP} and the basic information is summarized in Table \ref{tab:datasets}. To simulate the label skew distribution across subgraphs in the subgraph-FL, we employ \(\alpha\)-based Dirichlet distribution to create imbalanced label distributions across clients, where \(\alpha\) controls the concentration of probabilities across label classes. 

\noindent \textbf{Baseline}. We compare the proposed LLM4FGL with seven baselines including two common FL algorithm (FedAvg \cite{DBLP:journals/corr/McMahanMRA16}, FedProx \cite{DBLP:conf/mlsys/LiSZSTS20}) and five subgraph-FL algorithms (FedSage+ \cite{DBLP:conf/nips/ZhangYLSY21}, Fed-PUB \cite{DBLP:conf/icml/BaekJJYH23}, FedGTA \cite{DBLP:journals/corr/abs-2401-11755},  AdaFGL \cite{DBLP:conf/icde/LiWZSLW24}, FedTAD \cite{DBLP:conf/ijcai/ZhuLWWHL24}). These baselines are implemented within the OpenFGL \cite{DBLP:journals/corr/abs-2408-16288} framework to ensure consistency in evaluation and comparison. The details of the baselines are summarized as follows.

\begin{itemize}[leftmargin=10pt]
\item FedAvg \cite{DBLP:journals/corr/McMahanMRA16} serves as a foundation method in FL, performing model aggregation by averaging the updates from individual clients.

\item FedProx \cite{DBLP:conf/mlsys/LiSZSTS20} enhances FedAvg by introducing a proximal term to address systems and statistical heterogeneity.

\item FedSage+ \cite{DBLP:conf/nips/ZhangYLSY21} trains missing neighborhood and feature generators to predict the graph structure and associated embeddings.

\item FedGTA \cite{DBLP:journals/corr/abs-2401-11755} is a personalized optimization strategy that optimizes through topology-aware local smoothing confidence and mixed neighbor features. 

\item Fed-PUB \cite{DBLP:conf/icml/BaekJJYH23} is a personalized fgl framework that addresses heterogeneity by using functional embedding-based similarity matching and weighted averaging to optimize local client.

\item AdaFGL \cite{DBLP:conf/icde/LiWZSLW24} proposes a decoupled two-stage personalized method, improves performance by adaptively combining propagation results in both homophilous and heterophilous data scenarios.

\item FedTAD \cite{DBLP:conf/ijcai/ZhuLWWHL24} addresses heterogeneity issues by using a topology-aware data-free knowledge distillation to enhance class-wise knowledge transfer between local and global models.
\end{itemize}

\noindent \textbf{Implementation}. We adopt Gemma-2-9B-it \cite{gemma_2024} as the LLM model. The number of layers for MLP is 4 and the hidden size is 512. For GCN, the number of layers is 2 and the hidden size is 256. We adopt the typical train/validation/test data splitting for Cora, Citeseer, and Pubmed \cite{Chen2023ExploringTP}. The $\alpha$ to control the Dirichlet distribution is set to 100 for all datasets. For LLM-based graph generation, the number of generated neighbors of each client is tuned from \{1, 5, 10, 15, 20, 25, 30\} and the $k$ in the reflection is tuned from \{5, 10, 15, 20, 25, 30\}. For edge predictor, the $k$ is tuned from \{50, 100, 150, 200, 250, 300, 350, 400\}. All the baselines are tuned based on the performance on the validation dataset.

\begin{table*}[t]
\caption{Overall performance of different methods on Three datasets. The best results are in bold and runner-up results are in underline.}\label{tab:main_perf}
\vspace{-0.1in}
\label{tab:main:overlap}
\centering
\fontsize{8.3pt}{10pt}\selectfont
\resizebox{\textwidth}{!}{
\renewcommand{\arraystretch}{1.3}
\begin{tabular}{lccccccccc}
\toprule
& \multicolumn{3}{c}{\bf Cora} & \multicolumn{3}{c}{\bf CiteSeer} & \multicolumn{3}{c}{\bf PubMed} \\
\cmidrule(l{2pt}r{2pt}){2-4} \cmidrule(l{2pt}r{2pt}){5-7} \cmidrule(l{2pt}r{2pt}){8-10}

\textbf{Methods} & \textbf{5 Clients} & \textbf{7 Clients} & \textbf{10 Clients} & \textbf{5 Clients} & \textbf{7 Clients} & \textbf{10 Clients} & \textbf{5 Clients} & \textbf{7 Clients} & \textbf{10 Clients} \\
\midrule

Fedavg   & 75.55 $\pm$ 0.25 & \underline{75.27 $\pm$ 0.77} & \underline{74.61 $\pm$ 0.70} & 71.13 $\pm$ 0.40 & 70.22 $\pm$ 0.65 & 70.44 $\pm$ 0.83 & 77.37 $\pm$ 0.84 & 76.91 $\pm$ 0.29 & 78.50 $\pm$ 0.74 \\
FedProx & 76.85 $\pm$ 0.16 & 74.94 $\pm$ 0.74 & 74.18 $\pm$ 1.10 & \underline{72.42 $\pm$ 0.14} & \textbf{70.87} $\pm$ \textbf{0.45} & \underline{70.98 $\pm$ 0.36} & 78.48 $\pm$ 0.24 & 77.17 $\pm$ 0.40 & \underline{78.86 $\pm$ 0.12} \\ 
Fedsage+  & 75.98 $\pm$ 0.17 & 74.65 $\pm$ 0.43 & 74.30 $\pm$ 0.20 & 71.35 $\pm$ 0.24 & 69.71 $\pm$ 0.27 & 70.90 $\pm$ 0.77 & 78.52 $\pm$ 0.20 & 76.92 $\pm$ 0.19 & 78.10 $\pm$ 0.23 \\
FedPub & 73.50 $\pm$ 0.32 & 71.57 $\pm$ 0.54 & 66.27 $\pm$ 0.58 & 68.43 $\pm$ 0.25 & 65.03 $\pm$ 0.45 & 65.47 $\pm$ 0.30 & 71.11 $\pm$ 1.24 & 69.09 $\pm$ 0.55 & 70.56 $\pm$ 0.86 \\
FedGTA  & 75.36 $\pm$ 0.38 & 73.48 $\pm$ 0.06 & 72.96 $\pm$ 0.32 & 69.14 $\pm$ 1.19 & 68.61 $\pm$ 0.11 & 68.04 $\pm$ 0.44 & 77.17 $\pm$ 0.13 & 75.85 $\pm$ 0.40 & 78.35 $\pm$ 0.07 \\
FedTAD & \underline{76.96 $\pm$ 0.58} & 74.52 $\pm$ 0.44 & 73.49 $\pm$ 0.52 & 71.01 $\pm$ 0.19 & 69.86 $\pm$ 
0.21 & 70.57 $\pm$ 0.29 & \underline{78.75 $\pm$ 0.25} & \underline{77.34 $\pm$ 0.53} & 78.73 $\pm$ 0.26  \\
AdaFGL  & 75.20 $\pm$ 0.36 & 72.31 $\pm$ 0.98 & 72.89 $\pm$ 0.34 & 68.86 $\pm$ 0.28 & 67.48 $\pm$ 0.32 & 68.58 $\pm$ 0.81 & 76.97 $\pm$ 0.33 & 75.61 $\pm$ 0.13 & 78.12	$\pm$ 0.11  \\

\midrule

LLM4FGL (Ours)    & \textbf{77.49} $\pm$ \textbf{0.20} & \textbf{77.30} $\pm$ \textbf{0.34} & \textbf{77.83} $\pm$ \textbf{0.56} & \textbf{72.64} $\pm$ \textbf{0.05} & \underline{70.85 $\pm$ 0.47} & \textbf{72.10} $\pm$ \textbf{0.24} & \textbf{80.23} $\pm$ \textbf{0.10} & \textbf{79.10} $\pm$ \textbf{0.10} & \textbf{80.87} $\pm$ \textbf{0.01} \\

\bottomrule

\end{tabular}
}
\end{table*}

\begin{table*}[t]
\caption{Performance of the LLM4FGL as a plug-in on baselines.}
\vspace{-0.1in}
\label{tab:plug_in}
\small
\centering
\resizebox{\textwidth}{!}{
\renewcommand{\arraystretch}{1.0}
\begin{tabular}{cc|ccccccc}
\toprule

\textbf{Datasets} & \textbf{Method} & \textbf{Fedavg}  & \textbf{FedProx} & \textbf{Fedsage+} & \textbf{FedPub} & \textbf{FedGTA} & \textbf{FedTAD} & \textbf{AdaFGL} \\
\midrule 
\multirow{2}{*}{Cora} 
&original   & 74.61$\pm$ 0.70 & 74.18 $\pm$ 1.10 & 74.30 $\pm$ 0.20 & 66.27 $\pm$ 0.58 & 72.96 $\pm$ 0.32 & 73.49 $\pm$ 0.52 & 72.89 $\pm$ 0.34  \\
&+LLM4FGL  & 77.83 $\pm$ 0.56 & 76.87 $\pm$ 0.94 & 75.91 $\pm$ 0.67 & 70.54 $\pm$ 0.33 & 77.40 $\pm$ 0.67 & 76.98 $\pm$ 0.23 & 73.12 $\pm$ 0.65  \\
\midrule
\multirow{2}{*}{CiteSeer} 
&original   & 70.44 $\pm$ 0.83 & 70.98 $\pm$ 0.36 & 70.90 $\pm$ 0.77 & 65.47 $\pm$ 0.30 & 68.04 $\pm$ 0.44 & 70.57 $\pm$ 0.29 & 68.58 $\pm$ 0.81 \\
&+LLM4FGL  & 72.10 $\pm$ 0.24 & 71.76 $\pm$ 0.26 & 70.87 $\pm$ 0.12 & 68.56 $\pm$ 0.19 & 70.13 $\pm$ 0.45 & 72.98 $\pm$ 0.25 & 71.12 $\pm$ 0.34  \\
\midrule
\multirow{2}{*}{PubMed} 
&original   & 78.50 $\pm$ 0.74 & 78.86 $\pm$ 0.12 & 78.10 $\pm$ 0.23 & 70.56 $\pm$ 0.86 & 78.35 $\pm$ 0.07 & 78.73 $\pm$ 0.26 & 78.12 $\pm$ 0.11  \\
&+LLM4FGL  & 78.87 $\pm$ 0.01 & 79.23 $\pm$ 0.56 & 79.11 $\pm$ 0.24 & 72.98 $\pm$ 0.34 & 80.01 $\pm$ 0.56 & 79.98 $\pm$ 0.09 & 78.92 $\pm$ 0.34  \\
\bottomrule

\end{tabular}
}
\end{table*}

\subsection{Performance against baselines (RQ1)}
To answer RQ1, we set the number of federated clients to 5, 7, and 10, respectively, and report the performance of different methods on three datasets. As reported in Table~\ref{tab:main_perf}, we make  the following observations:

(1) LLM4FGL consistently outperforms all baselines, achieving the SOTA performance. Specifically, LLM4FGL achieves average relative improvements of 2.57\%, 0.62\%, and 1.63\% on Cora, Citeseer, and PubMed over runner-up results. This verifies the effectiveness of LLM4FGL in mitigating the data heterogeneity problem on FGL. 

(2) FGL baselines generally have poorer performance compared to typical FL baselines (e.g., Fedavg and FedProx), especially when the number of clients is large. We employ the label imbalance splitting method to divide the whole graph and result in more missing cross-client edges compared to the traditional community-based graph partition algorithm (e.g., Louvain \cite{DBLP:journals/corr/abs-2311-06047}). However, these FGL baselines mostly focus on model-level designs, failing to tackle missing information at the data level. On the contrary, LLM4FGL directly augments local data using LLMs, achieving stable performance among different numbers of clients and largely improving the performance.

(3) Surprisingly, LLM4FGL even outperforms some personalized FGL methods, such as FedPub and FedGTA. By training different models for different clients, personalized FGL methods are specialized in tackling heterogeneity data distributions. The superiority of LLM4FGL demonstrates that the data-level augmentation not only largely alleviates the heterogeneity but also reduces local personalized computing.



\subsection{Performance as a plug-in (RQ2)}
As a flexible data pre-processing method, the proposed LLM4FGL can recover a complete local graph for each client, and the recovered graphs are fit for various federated graph models. By integrating with existing models that address data heterogeneity in FGL, LLM4FGL serves as a plug-in to further enhance them. We report the performance comparison of different baselines under the Original and integrated LLM4FGL in Table~\ref{tab:plug_in}. According to the results, we can find that LLM4FGL consistently improves the accuracy across all baselines in three datasets. For Cora, Citeseer, and Pubmed datasets, LLM4FGL on average brings 3.88\%, 2.81\%, and 1.62\% relative improvements in accuracy on seven baselines. 
Compared to Cora and CiteSeer, the gains on Pubmed are relatively smaller, attributed to the fact that Pubmed has rich textual features and a relatively dense structure, making the gains from additional information marginal. We can also find that by utilizing the augmented graph by LLM4FGL, some baselines can outperform LLM4FGL (i.e., with Fedavg as a backbone), demonstrating that simultaneously considering data-level and model-level strategies to tackle heterogeneity is significant.
%


\begin{table}[t]
\caption{Ablation study on key components under 10 clients.}
\label{tab:ab}
\centering
\begin{tabular}{l|ccc}
\toprule
\textbf{Method}   & \multicolumn{1}{c}{\bf Cora} & \multicolumn{1}{c}{\bf CiteSeer} & \multicolumn{1}{c}{\bf PubMed} \\

\midrule

Original   & 74.61 $\pm$ 0.70 & 70.44 $\pm$ 0.83 & 78.50 $\pm$ 0.74  \\
\midrule
w/o R  & 77.48 $\pm$ 0.34 & 71.54 $\pm$ 0.25 & 78.94 $\pm$ 0.34  \\
w/o E & 77.66 $\pm$ 0.23 & 72.01 $\pm$ 0.33 & 80.81 $\pm$ 0.14  \\
w/o R\&E & 77.10 $\pm$ 0.12 & 71.56 $\pm$ 0.12 & 78.87 $\pm$ 0.41  \\

\midrule
LLM4FGL  & \textbf{77.83 $\pm$ 0.56} & \textbf{72.11 $\pm$ 0.24} & \textbf{80.87 $\pm$ 0.01}  \\

\bottomrule

\end{tabular}
\end{table}

\subsection{Ablation study (RQ3)}

\makeatletter
\renewcommand{\thesubfigure}{\fontsize{10pt}{12pt}\selectfont (\alph{subfigure})}
\makeatother
\begin{figure*}[t]
\centering
\subfigure[\fontsize{10pt}{12pt}\selectfont \textrm{top-k of edge predictor}]
{
 	\begin{minipage}[b]{.25\linewidth}
        \centering
        \includegraphics[width=\textwidth]{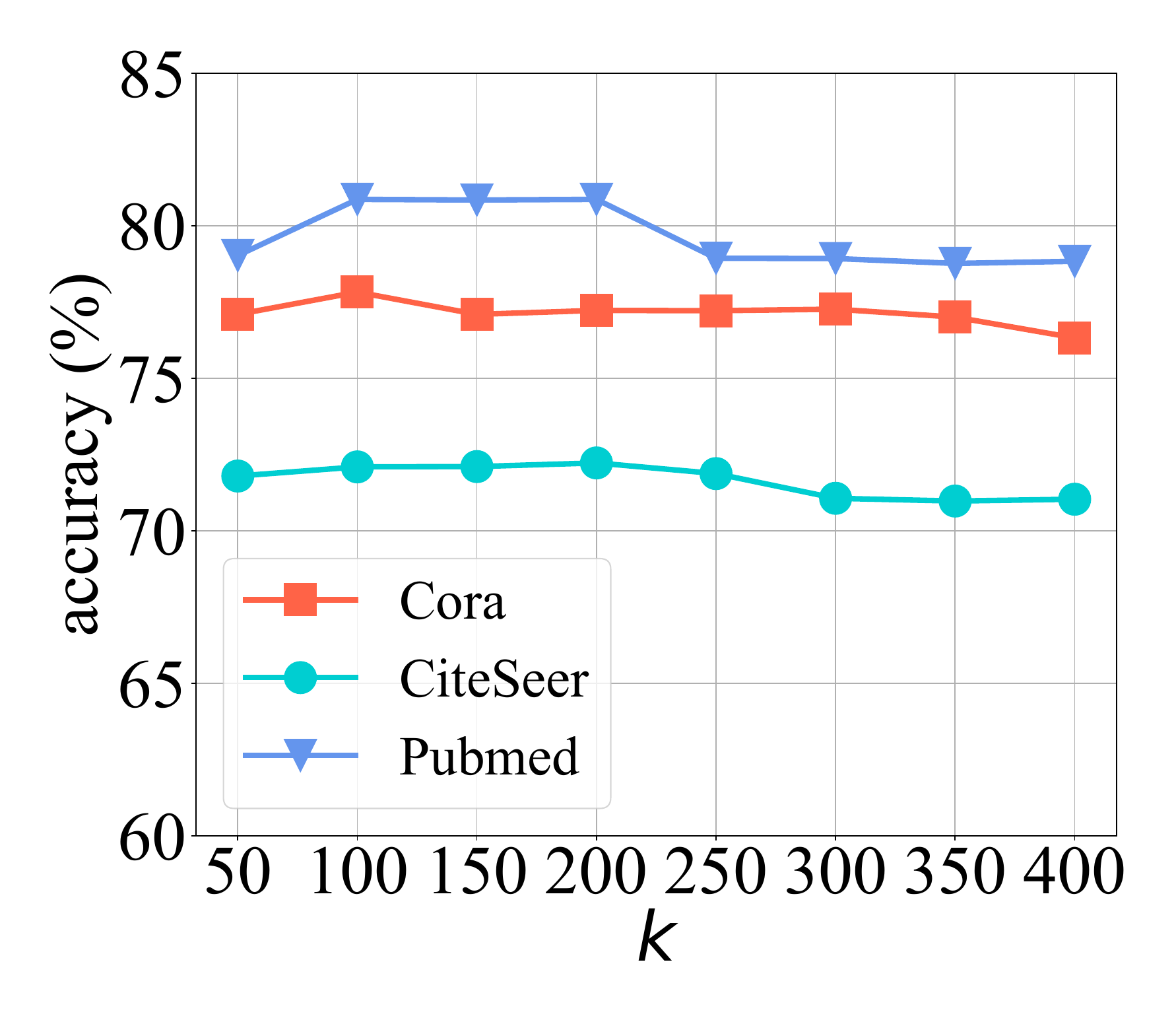}
        \vspace{-20pt}
    \end{minipage}
}
\subfigure[\fontsize{10pt}{12pt}\selectfont \textrm{top-k of reflection}]
{
    \begin{minipage}[b]{.25\linewidth}
        \centering
        \includegraphics[width=\textwidth]{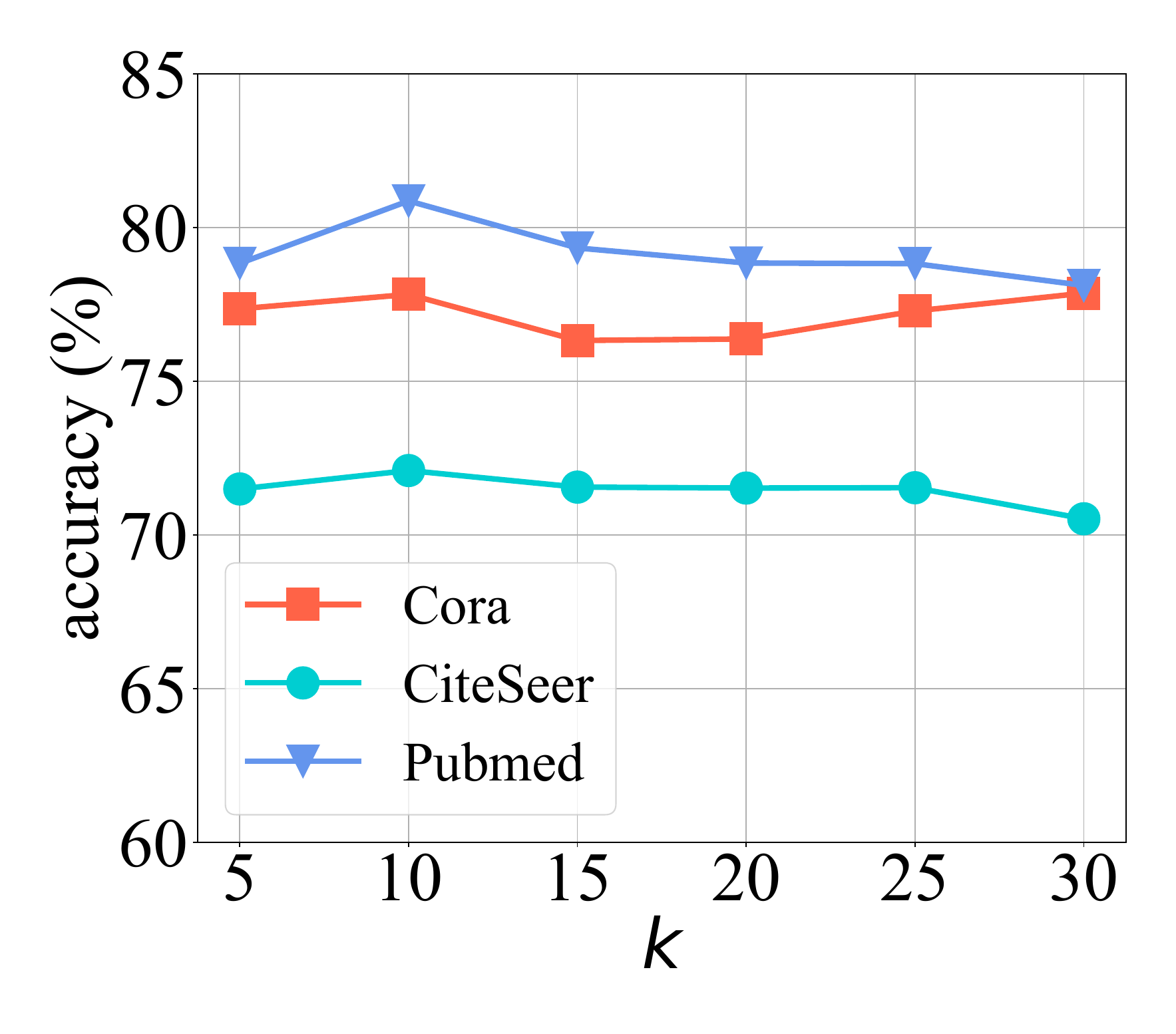}
        \vspace{-20pt}
    \end{minipage}
}
\subfigure[\fontsize{10pt}{12pt}\selectfont \textrm{generated node number n}]
{
 	\begin{minipage}[b]{.25\linewidth}
        \centering
        \includegraphics[width=\textwidth]{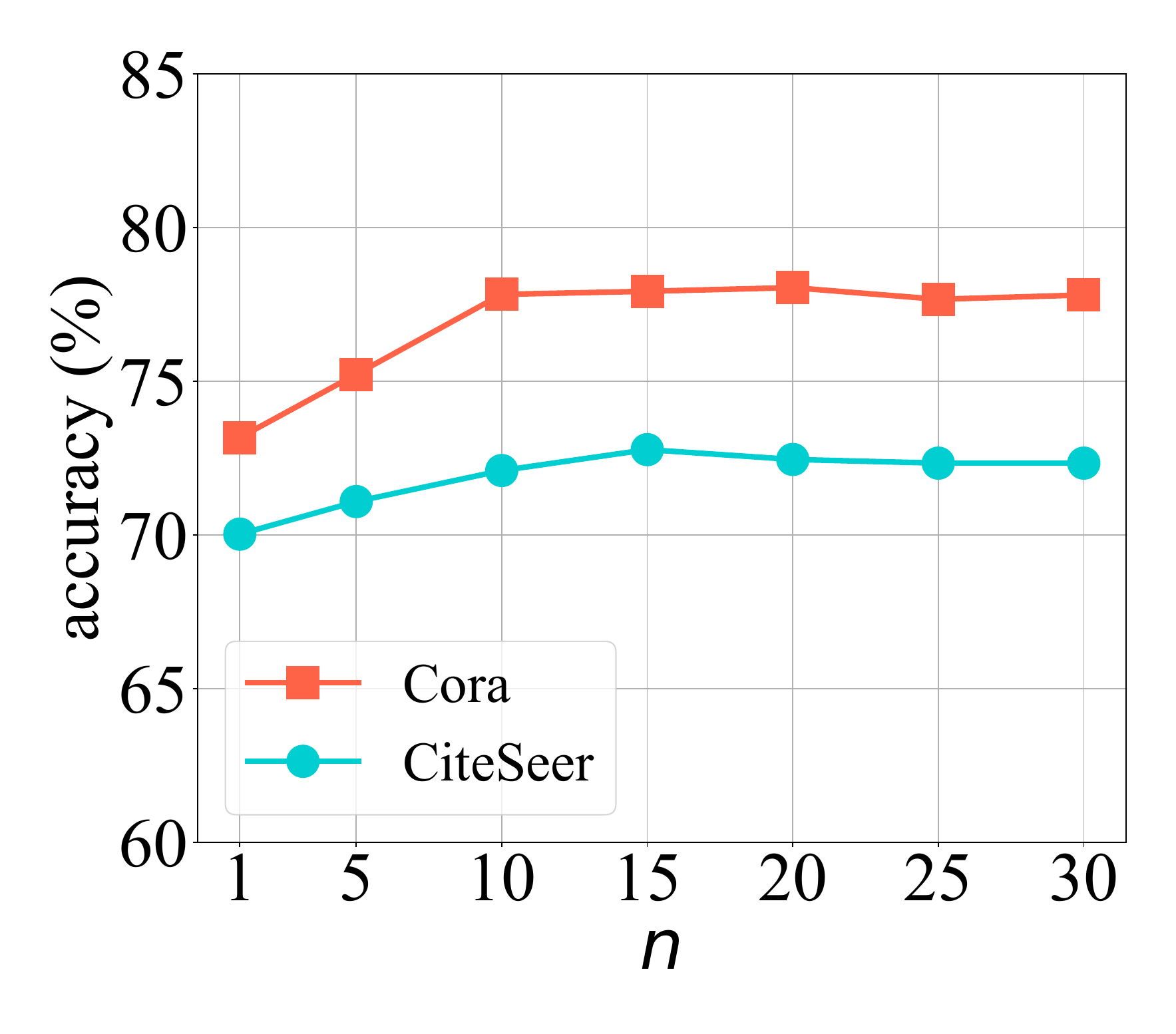}
        \vspace{-20pt}
    \end{minipage}
}
    \vspace{-0.05in}
    \caption{Effect of $k$ (top-$k$ selection) and $n$ (number of generated nodes for one node) on model performance.}
    \label{fig: param}
\end{figure*}

\begin{figure*}[tp]
    \centering
    \includegraphics[width=0.98\linewidth]{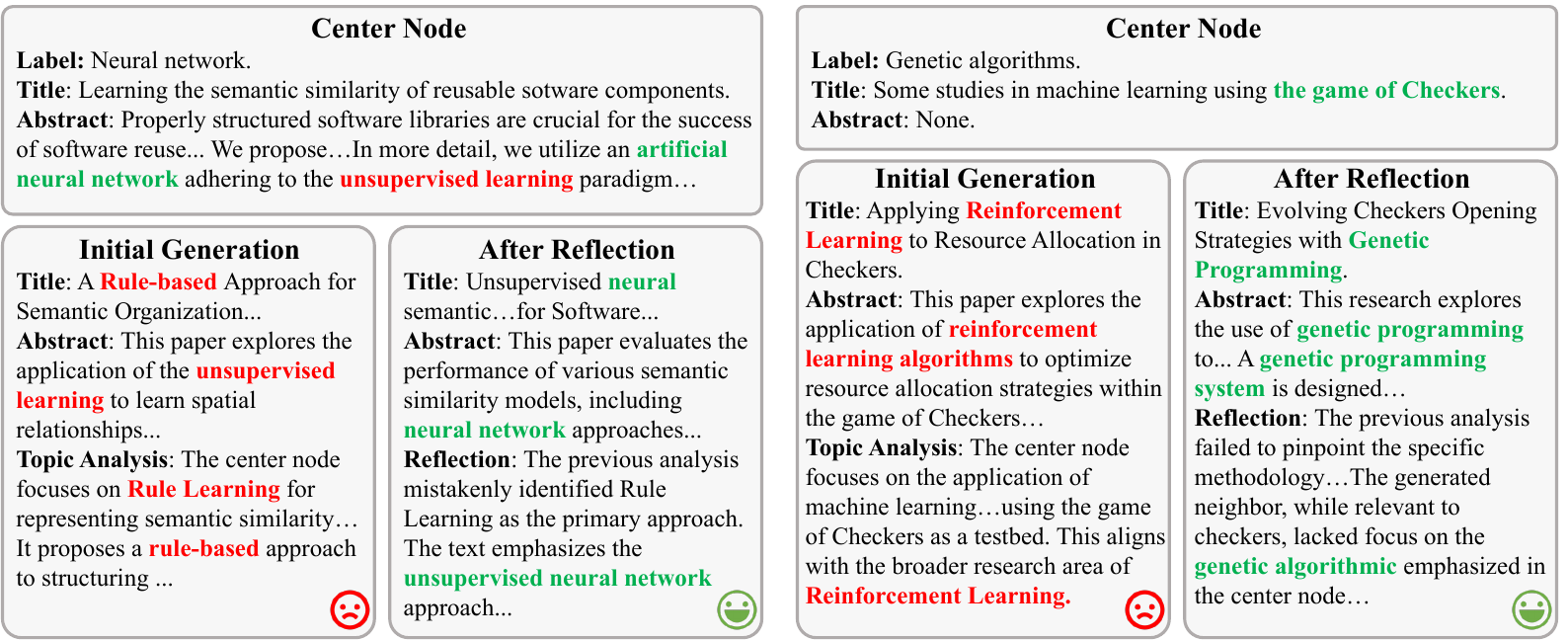}
    \caption{Case study of generated nodes on Cora dataset.}
    \label{fig:case}
    \vspace{-0.05in}
\end{figure*}

To answer RQ3, we focus on two key components in training LLM4FGL: (1) a generation-and-reflection mechanism specifically designed for LLMs, and (2) an edge predictor used to predict missing structures. We conduct the ablation study on three datasets (Cora, CiteSeer, Pubmed) with 10 participating clients to evaluate the impact of two modules. The experimental result is presented in Table \ref{tab:ab}, where \textit{w/o R} represents removing the reflection mechanism, \textit{w/o E} represents removing the edge predictor, and \textit{w/o R\&E} represents removing both of them. We also report vanilla FGL (\textit{original}) without LLM-based generation for comparison. As observed, both reflection and edge predictor contribute to the overall performance. The reflection mechanism plays a more important role than the edge predictor in improving the generated textural quality, especially in PubMed dataset. In our framework, the generated textural neighbor will connect to the center node and directly affect the center node's category. In other words, the edge predictor operating on high-quality generated nodes will further improve the performance but on low-quality nodes (without reflection) have little effect. We can also see that the LLM-based generation brings the largest gains (w/o R\&E can also significantly improve vanilla FGL), which demonstrates the great potential of LLMs for tackling the heterogeneity issues in FGL).

\subsection{Hyperparameter analysis (RQ4)}
To further validate how the hyperparameters impact the performance of LLM4FGL, we conduct hyperparameter sensitivity analysis experiments on the three datasets. Specifically, we first present the accuracy of LLM4FGL under different $k$ in the edge predictor in Figure~\ref{fig: param} (a). The results indicate that the accuracy of LLM4FGL will first increase and then decay along with the increases of $k$, indicating a larger $k$ may introduce more noise to the local structure and a smaller $k$ may not be enough to obtain optimal structure. Besides, we report the accuracy of LLM4FGL under different numbers $k$ of reflection nodes in one client in Figure~\ref{fig: param} (b), and find the same trend. This is because a smaller $k$ may not fully consider all nodes with low confidence, leading to a low generated quality, and a larger $k$ will involve nodes with true prediction thus undermining the accuracy. Lastly, we show the influence of generated neighbor number $n$ on LLM4FGL's performance in Figure~\ref{fig: param} (c), and the results indicate that a proper $n$ can significantly improve the performance but the gains become smaller along with the increase of $n$. It demonstrates that the key factor to improve the performance is not the quantity of additional information but the quality of that, which also matches the insights of our designed reflection mechanism. Smaller $n$ will significantly weaken the performance as the smaller $n$ make the model vulnerable to generated noise nodes.


\subsection{Case study}
Finally, we conduct a case study to show the generated node text by LLMs. Figure \ref{fig:case} presents two cases from the Cora dataset. In the first case, the center node's true topic is \textit{neural network}, but LLMs have a wrong focus on text \textit{unsupervised learning}. As a result, LLM generates node text with \textit{rule learning} topic. After reflection, LLM catches the key text \textit{neural network} and regenerates a node with a neural network topic. In the second case, the center node has incomplete information (lacking the abstract of the paper). It's difficult for LLM to correctly identify the topic $\textit{Genetic Algorithms}$ solely based on the title. Furthermore, owning the knowledge that $\textit{Reinforcement Learning}$ can be utilized to simulate $\textit{game of Checkers}$, LLM misunderstands it as $\textit{reinforcement learning}$. After reflection, LLM gives the true topic considering that genetic algorithms can also be used to design strategies. The above two cases demonstrate that the reflection mechanism improves the quality of generated data and thus the center node's information can be well complemented.

\section{Conclusion}

In this work, we first explore the potential of LLMs in federated graph learning. Specifically, we leverage the powerful text understanding and generation capabilities of LLMs to address the heterogeneity problem in FGL from the data perspective. To achieve this, we propose LLM4FGL, a framework that decomposes the LLM-based generation task into two distinct sub-tasks theoretically: locally utilizing LLM to generate missing neighbors, and jointly pre-training an edge predictor across all clients to infer connections between generated and raw nodes. Experiments on three real-world datasets demonstrate the effectiveness of our approach. Besides, as a plug-in, our framework can also significantly improve existing FGL methods.


\clearpage
\bibliographystyle{ACM-Reference-Format}
\bibliography{sample-base}

\appendix

\end{document}